%% file: main.tex
\documentclass[letterpaper, 10 pt, conference]{ieeeconf}  

\IEEEoverridecommandlockouts                              
                                                          
\usepackage{amsmath, amssymb, graphicx, url, pgf, bm, subcaption, float}
\usepackage[ruled,vlined]{algorithm2e}
\graphicspath{{figures/}}

\usepackage[colorinlistoftodos, textsize=small, textwidth=50mm]{todonotes}
\definecolor{lfcolor}{rgb}{.7, 0.1, 0.7}

\definecolor{bhcolor}{rgb}{.0, .0, .0}

\newcommand{\bhr}[2]{#2}
\renewcommand{\bhr}[2]{{\color{bhcolor}{#1}}}

\newcommand{\nls}{\bm{x}}
\newcommand{\s}{\bm{s}}
\newcommand{\pixel}{\bm{p}}
\newcommand{\obs}{\bm{o}}

\newcommand{\cost}{c}
\newcommand{\act}{\bm{a}}
\newcommand{\dact}{\Delta\act}
\newcommand{\eig}{\varphi}
\newcommand{\OU}{\bm{\epsilon}}
\newcommand{\policy}{\pi}

\newcommand{\obsh}{\hat{\obs}}

\newcommand{\inv}{\textsf{-1}}
\newcommand{\tr}{\textsf{T}}
\DeclareMathOperator{\iu}{{i\mkern1mu}}

\newcommand{\pde}[1][]{\ensuremath{\bigtriangledown_{#1}}}
\newcommand{\ddt}[1][]{\ensuremath{\frac{\text{d}#1}{\text{d}t}}}
\newcommand{\loss}[1]{\mathcal{L}_\text{#1}}

\newcommand{\idx}[1]{\ensuremath{{[#1]}}}

\newcommand{\inR}[1]{\in \mathbb{R}^{#1}}
\newcommand{\norm}[1]{\left\lVert#1\right\rVert}
\SetKwComment{Comment}{$\triangleright$\ }{}
\newcommand{\citep}[1]{\cite{#1}}
\newcommand{\citet}[1]{\cite{#1}}
\newcommand{\figureref}[1]{Fig.~\ref{#1}}
\renewcommand{\eqref}[1]{(\ref{#1})}
\newcommand{\algorithmref}[1]{Algorithm~\ref{#1}}
\newcommand{\sectionref}[1]{Section~\ref{#1}}
\title{DeepKoCo: Efficient latent planning with a \bhr{task-relevant}{robust} Koopman representation}
\usepackage{times}

\author{Bas van der Heijden$^{1}$, Laura Ferranti$^{1}$, Jens Kober$^{1}$, Robert Babu{\v{s}}ka$^{1}$
\thanks{*This work was supported by the European Union's H2020 project Open Deep Learning Toolkit for Robotics (OpenDR) under grant agreement No 871449.}
\thanks{$^{1}$Cognitive  Robotics at the Faculty  of  3mE, Delft  University  of  Technology, The Netherlands. {\tt\small d.s.vanderheijden@tudelft.nl}}%
}

\begin{document}

\maketitle

\begin{abstract}%
This paper presents DeepKoCo, a novel model-based \bhr{}{reinforcement learning }agent that learns a latent Koopman representation from images. This representation allows DeepKoCo to plan efficiently using linear control methods, such as linear model predictive control. Compared to traditional agents, DeepKoCo \bhr{learns task-relevant dynamics}{is robust to task-irrelevant dynamics}, thanks to the use of a tailored lossy autoencoder network that allows DeepKoCo to learn latent dynamics that reconstruct and predict only observed costs, rather than all observed dynamics. As our results show, DeepKoCo achieves similar final performance as traditional model-free methods on complex control tasks while being considerably more robust to distractor dynamics, making the proposed agent more amenable for real-life applications.
\end{abstract}


\input{introduction.tex}
\input{related_work.tex}
\input{preliminaries.tex}
\input{latent_model.tex}
\input{mpc.tex}
\input{results.tex}
\input{future_work.tex}

\bibliographystyle{IEEEtran}
\bibliography{references}

\end{document}

%% file: introduction.tex
\section{INTRODUCTION}
\label{sec:intro}
\bhr{From self-driving cars to vision-based robotic manipulation, emerging technologies are characterized by visual measurements of strongly nonlinear physical systems. Unlike in highly controlled lab environments where any measured change is likely relevant, cameras in real-world settings are notorious for mainly capturing task-irrelevant information, such as, the movement of other robots outside of a manipulator's workspace or cloud movements captured by the cameras of self-driving cars.

While Deep Reinforcement Learning (DRL) algorithms can learn to perform various tasks using raw images, they will require an enormous number of trials. Prior methods mitigate this by encoding the raw images into a lower-dimensional representation that allows for faster learning. However, these methods can be easily distracted by irrelevant dynamics \cite{Zhang2020a}. This motivates data-driven methodologies that learn low-dimensional latent dynamics that are task-relevant and useful for control.}{From self-driving cars to vision-based robotic manipulation, emerging technologies are characterized by high-dimensional measurements of (strongly) nonlinear physical systems. Oftentimes, these systems are complex and lack simple models suitable for control design. Moreover, the high-dimensional measurements are typically contaminated with irrelevant dynamics that have nothing to do with the control objective. This motivates data-driven methodologies that learn low-dimensional latent dynamics that are robust to task-irrelevant dynamics and useful for control.} 

In the learning of latent dynamics for control, there is a trade-off between having an accurate dynamic model and one that is suitable for control. On one hand, latent dynamic models based on neural networks (NN) can provide accurate predictions over long horizons. On the other hand, their inherent nonlinearity renders them incompatible with efficient planning algorithms. Alternatively, one can choose to approximate the latent dynamics with a more restricted function approximation class to favor the use of efficient planning algorithms. In this respect, a promising strategy is represented by the Koopman framework \citep{Kaiser2017}. Loosely speaking, this framework allows one to map observations with nonlinear dynamics to a latent space where the \textit{global dynamics} of the autonomous system are approximately linear (Koopman representation). This enables the use of powerful linear optimal control techniques in the latent space \citet{Kaiser2017}.

While the Koopman framework is promising, existing methods have fundamental limitations that must be addressed to fully exploit the benefits of this method for control applications. First, methods that identify Koopman representations from data were designed for prediction and estimation. These methods were later adapted for control. These adaptations, however, lead to limiting assumptions on the underlying dynamics\bhr{, such as assuming the Koopman representation to be linear in the states and actions \citep{Arbabi2018a, Mamakoukas2019, Bruder2019, Li2020a, Korda2020a}}{}. Second, the\bhr{se}{ existing} methods are \emph{task agnostic}, that is, the models represent all dynamics they observe, whether they are relevant to the task or not\bhr{. This focuses the majority of their model capacity on potentially task-irrelevant dynamics.}{ \citep{Zhang2020a}} 

Therefore, we introduce \emph{Deep Koopman Control} (DeepKoCo), that is, a model-based \bhr{}{reinforcement learning }agent that learns a latent Koopman representation from raw pixel images and achieves its goal through planning in this latent space. The representation is \emph{(i)} robust to task-irrelevant dynamics and \emph{(ii)} compatible with efficient planning algorithms. We propose a lossy autoencoder network that reconstructs and predicts observed costs, rather than all observed dynamics, which leads to a representation that is \bhr{task-relevant.}{robust to task-irrelevant dynamics.} The latent-dynamics model can represent continuously differentiable nonlinear systems and does not require knowledge of the underlying environment dynamics or cost function. We demonstrate the success of our approach on two continuous control tasks and show that our method is more robust to irrelevant dynamics than state-of-the-art approaches, that is, DeepMDP \citep{Gelada2019a} and Deep Bisimulation for Control (DBC) \citep{Zhang2020a}.

%% file: related_work.tex
\section{RELATED WORK}
\subsubsection{Koopman control} Koopman theory has been used to control various nonlinear systems with linear control techniques, both in simulation \citep{Kaiser2017, Korda2018, Li2020a} and in real-world robotic applications \citep{Mamakoukas2019, Bruder2019}. Herein, \citet{Brunton2016a, Kaiser2017, Mamakoukas2019} used a linear quadratic regulator (LQR), while \citet{Korda2018, Arbabi2018a, Bruder2019, Li2020a, Korda2020a} applied linear model predictive control (MPC). \citet{Korda2018, Arbabi2018a, Bruder2019, Korda2020a} used data-driven methods that were derived from the Extended Dynamic Mode Decomposition (EDMD) \citep{Williams2015} to find the Koopman representation. In contrast, \citet{Mamakoukas2019, Kaiser2017, Brunton2016a} require prior knowledge of the system dynamics to hand-craft parts of the lifting function. Similar to \citet{Li2020a}, we rely on deep learning to derive the Koopman representation for control. However, we do not assume the Koopman representation to be (bi-)linear in the states and actions and we show how our representation can be used to control systems that violate this assumption. Compared to existing methods, we propose an agent that learns the representation online in a reinforcement learning setting using high-dimensional observations that contain irrelevant dynamics.

\subsubsection{Latent planning} Extensive work has been conducted to learn latent dynamics from images and use them to plan suitable actions \citep{Watter2015a, Banijamali2018a, Hafner2019a}. \citet{Hafner2019a} proposes a model-based agent that uses NNs for the latent dynamics and cost model. To find suitable action sequences, however, their method requires a significant computational budget to evaluate many candidate sequences. Alternatively, \citet{Watter2015a, Banijamali2018a} propose locally linear dynamic models, which allowed them to efficiently plan for actions using LQR. However, their cost function was defined in the latent space and required observations of the goal to be available. In contrast to our approach, all aforementioned methods are trained towards full observation reconstruction, which focuses the majority of their model capacity on potentially task-irrelevant dynamics.

\subsubsection{Relevant representation learning} \citet{Gelada2019a, Zhang2020a} filter task-irrelevant dynamics by minimizing an auxiliary bisimulation loss. Similar to our approach, they propose learning latent dynamics and predicting costs. Their method, however, is limited to minimizing a single-step prediction loss, while we incorporate multi-step predictions. This optimizes our model towards accurate long-term predictions. \citet{Oh2017a, Schrittwieser2019a} also proposed training a dynamics model towards predicting the future sum of costs given an action sequence. However, their method focused on discrete control variables, while we focus on continuous ones.

%% file: preliminaries.tex
\section{PRELIMINARIES}
This section \bhr{briefly introduces}{presents} the Koopman framework for autonomous and controlled nonlinear systems. \bhr{A detailed description can be found in \citep{Kaiser2017}. }{}This framework is fundamental to the design of our latent model and control strategy. \looseness=-1
\subsection{Koopman eigenfunctions for autonomous systems}
Consider the following autonomous nonlinear system $\dot{\obs} = F(\obs)$, where the observations $\obs\inR{N}$ evolve according to the smooth continuous-time dynamics $F(\obs)$. For such a system, there exists a lifting function $g(\cdot): \mathbb{R}^{N} \to \mathbb{R}^{n}$ that maps the observations to a latent space where the dynamics are linear, that is,
\begin{align}\label{eq:Koopman_general}
	\ddt g(\obs) = \bhr{\mathcal{K}}{K} \circ g(\obs)\bhr{,}{.}
\end{align}
\bhr{where $\mathcal{K}$ is the infinitesimal operator generator of Koopman operators $K$. In theory, $K$ is infinite dimensional (i.e., $n \to {\infty}$), but a finite-dimensional matrix representation can be obtained by restricting it to an invariant subspace. Any set of eigenfunctions of the Koopman operator spans such a subspace. Identifying these eigenfunctions \citep{Korda2018, Lusch2018}  provides a set of intrinsic coordinates that enable global linear representations of the underlying nonlinear system. A Koopman eigenfunction satisfies}{Loosely speaking, the \bhr{infinitesimal }{}Koopman operator \bhr{generator $\mathcal{K}$}{$K$} acts as a matrix on the lifting function $g(\cdot)$. With a slight abuse of notation, we will write it as a matrix multiplication in the sequel. In theory, a latent space with infinite dimension \bhr{could be}{is} required for an exact solution (i.e., $n \to {\infty}$). Fortunately, a finite approximation (with $n$ sufficiently large) provides a latent representation where the global dynamics are approximately linear. Direct identification of the lifting function and Koopman operator with the formulation in \eqref{eq:Koopman_general} often leads to inaccurate representations, so \citep{Korda2018, Lusch2018} seek to identify eigenfunctions of the Koopman operator instead, defined as}
\begin{align}
	\ddt \phi(\obs) &= K \phi(\obs) = \lambda \phi(\obs), \label{eq:eigenfunction}
\end{align}
where $\lambda\bhr{\in \mathbb{C}}{}$ is the continuous-time eigenvalue corresponding to eigenfunction $\phi(\obs)$.

\subsection{Koopman eigenfunctions for controlled systems}
For controlled nonlinear system \bhr{}{$\obs = \tilde{F}(\obs, \act)$}with action $\act\bhr{\inR{m}}{}$ and smooth continuous-time dynamics $\dot{\obs} = \tilde{F}(\obs, \act)$\bhr{}{and action $\act$}, we follow the procedure in \cite{Kaiser2017}. Given the eigenfunction $\phi(\obs, \act)$ augmented with $\act$ for the controlled system, we can take its time derivative and apply the chain rule with respect to $\obs$ and $\act$, leading to
\begin{align}\label{eq:Koopman_control_cont}
	\ddt\phi(\obs, \act) &= \underbrace{\pde[\obs] \phi(\obs, \act)\tilde{F}(\obs, \act)}_{\lambda \phi(\obs, \act)} + \pde[\act]\phi(\obs, \act) \dot{\act},
\end{align}
where $\lambda$ is now the eigenvalue that corresponds to eigenfunction $\phi(\obs, \act)$. \bhr{Since $\dot{\act}$ can be chosen arbitrarily, we could set it to zero and instead interpret each action as a parameter of the Koopman eigenfunctions. Thus, for any given choice of parameter $\act$ the standard relationship in \eqref{eq:eigenfunction} is recovered in the presence of actions. A local approximation of the Koopman representation is obtained when $\dot{\act}$ is nonzero.}{This partial differential equation is parametrized by $\act$ and depends on the derivative $\dot{\act}$ which can be chosen arbitrarily.}

\subsection{Identifying Koopman eigenfunctions from data}
To facilitate eigenfunction identification with discrete data, \eqref{eq:Koopman_control_cont} can be discretized with a procedure similar to \citep{Lusch2018}. The eigenvalues $\lambda_{\pm}=\mu \pm \iu \omega$ are used to parametrize block-diagonal $\Lambda\bhr{=\text{diag}(J^\idx{1}, J^\idx{2},...,J^\idx{P})\inR{2P\times2P}}{(\mu,\omega)}$. For \bhr{all $P$ pairs}{each pair} of complex eigenvalues, the discrete-time operator $\Lambda$ has a Jordan real block of the form
\begin{align}
	J(\mu, \omega) &= e^{\mu \Delta t}\begin{bmatrix} \cos(\omega \Delta t) && -\sin(\omega \Delta t)\\ \sin(\omega \Delta t) && \cos(\omega \Delta t)\end{bmatrix} \label{eq:jordan_block},
\end{align}
with sampling time $\Delta t$. The ``forward Euler method'' provides a discrete approximation of the control matrix, so that \eqref{eq:Koopman_control_cont} can be discretized as
\begin{align}\label{eq:Koopman_control}
	\eig(\obs_{k+1}, \act_{k+1}) &= \Lambda \eig(\obs_{k}, \act_{k}) + \underbrace{\pde[\act_{k}]\eig(\obs_{k}, \act_{k})}_{B_{\eig_k}} \underbrace{\dot{\act}_k \Delta t}_{\dact_k}.
\end{align}
Herein, the stacked vector $\eig=(\bhr{\phi}{\varphi}^\idx{1}, \bhr{\phi}{\varphi}^\idx{2},...,\bhr{\phi}{\varphi}^\idx{P})$ comprises \bhr{a set of}{} $P$ eigenfunctions with $\bhr{\phi}{\varphi}^\idx{j}\inR{2}$ associated with complex eigenvalue pair $\lambda^\idx{j}_{\pm}$ and Jordan block $J^\idx{j}$.\bhr{ Subscript $k$ corresponds to discretized snapshots in time.}{} If we view the action increment $\dact_k=\act_{k+1} - \act_k$ in \eqref{eq:Koopman_control} as the controlled input instead, we obtain a discrete \textit{control-affine} Koopman eigenfunction formulation with \textit{linear autonomous} dynamics for the original \textit{non-control-affine nonlinear} system. In the next section, we show that \eqref{eq:Koopman_control} plays a central role in our latent model.

%% file: latent_model.tex
\section{LEARNING \bhr{TASK-RELEVANT}{ROBUST} KOOPMAN EIGENFUNCTIONS}
\label{sec:learning_invariant_koopman_eigenfunctions}
For efficient planning in the latent space, we propose to learn a latent dynamics model that uses Koopman eigenfunctions as its latent state. This section describes this model and how the Koopman eigenfunctions can be identified robustly, that is, in a way that the identified eigenfunctions remain unaffected by task-irrelevant dynamics that are expected to contaminate the observations.
\subsection{Koopman latent model}\label{sec:latent_model}
\begin{figure*}[t]
	\centering
    \def\svgwidth{\textwidth}
	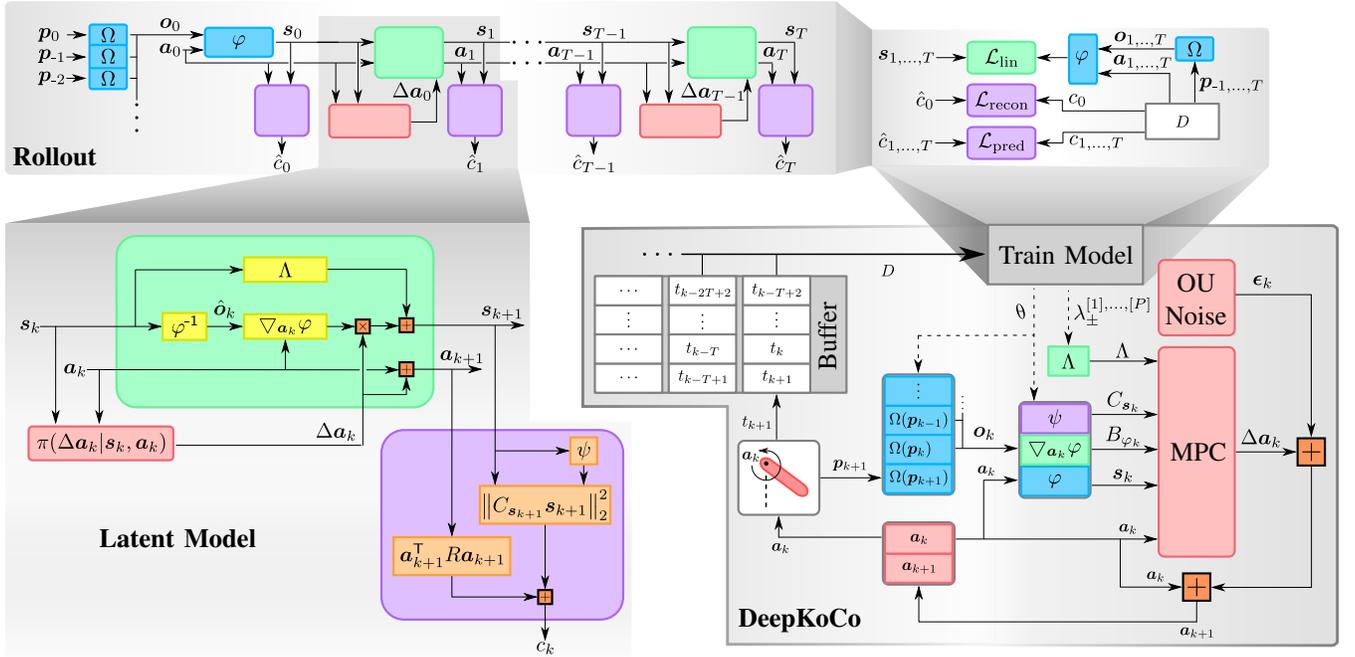
	\caption{\textbf{Latent Model} The proposed network architecture of the \emph{latent model}, consisting of the dynamic model, cost model, and policy (depicted in green, purple, and red, respectively). \textbf{Rollout} A multi-step ahead prediction with the \emph{latent model}. Note that we only encode an observation at the first time-step (blue boxes), after which we remain in the latent space. \textbf{DeepKoCo} The training procedure that corresponds to  \algorithmref{alg:deepKoCo}.}
	\label{fig:architecture}
\end{figure*}
We propose a \textit{lossy} autoencoder that builds on the deep autoencoder in \citet{Lusch2018}. Compared to \citet{Lusch2018}, our autoencoder enables control. To train the latent model, we provide the training objective that is to be minimized given a buffer $D$ that contains observed sequences $\{t_k\}^{T}_{k=0}$ of a Markov decision process with tuples $t_k=(\obs_k, \act_k, \dact_k, \cost_k)$, where $\cost_k$ are observed scalar costs. The proposed latent model is illustrated in \figureref{fig:architecture} with more details below on the individual components of the architecture.

\subsubsection{Encoder} The encoder $\eig$ is the \bhr{approximate }{}eigenfunction \bhr{}{transformation }that maps an observation-action pair $(\obs_k, \act_k)$ to the latent state $\s_k$. The encoder $\eig$ is parametrized by a neural network, defined as
\begin{align}
    \s_k &= \eig(\obs_{k}, \act_{k}). \label{eq:encoder}
\end{align}

\subsubsection{Latent dynamics} The latent state $\s_k$ \bhr{approximates}{is} a Koopman eigenfunction, so the autonomous time evolution in the latent space is linear and dictated by $\Lambda$. Note here that $\act_k$ is part of the \textit{augmented} latent state and we view the action increment $\dact_k$ as the controlled variable that is determined by the policy. This leads to the dynamics model in \eqref{eq:dynamics_model}, which we derived from \eqref{eq:Koopman_control}. The Koopman operator $\Lambda$ is parametrized by $P$ complex conjugate eigenvalue pairs $\lambda^\idx{j}_{\pm}$. We do not assume the latent dynamics to be linear in the control. Instead, the influence of $\dact_k$ on the latent state varies and depends on the partial derivative of the encoder with respect to the action, i.e., the state-dependent matrix $B_{\eig_k}=\pde[\act_{k}]\eig(\obs_{k}, \act_{k})\bhr{\inR{2P\times m}}{}$.
\begin{align}
	\begin{bmatrix} \s_{k+1} \\ \act_{k+1} \end{bmatrix} &= \begin{bmatrix} \Lambda & 0 \\ 0 & I \end{bmatrix} \begin{bmatrix} \s_k \\ \act_k \end{bmatrix} + \begin{bmatrix} B_{\eig_k} \\ I \end{bmatrix} \dact_k. \label{eq:dynamics_model}
\end{align}
\subsubsection{Cost model}
The environment contains a cost function that produces a scalar cost observation $\cost_k$ at every time-step. For planning in the latent space, we require a cost model $\hat{\cost}_k$  as a function of the latent state. This cost approximates the observed cost (i.e., $\hat{\cost}_k\! \approx\! \cost_k$). We adopt a latent state-dependent quadratic cost model to facilitate the use of fast planning algorithms (\sectionref{sec:control}). The \bhr{entries}{elements} of $C_{\s_k}\bhr{\inR{1\times2P}}{}$ are determined by a function $\psi(\s_k)$ that is parametrized by a neural network. The weights of $\psi$ are initially unknown and must be learned together with the rest of the latent model. We assume that the cost of applying action $\act_k$ is known \textit{a priori} and defined by matrix $R$. This leads to the cost model
\begin{align}
	\hat{\cost}_k &=  ||C_{\s_k}\s_k||^2_2 + \act^\tr_k R \act_k. \label{eq:cost_model}
\end{align}
\subsubsection{Policy} The action increment $\dact_k$ is the controlled variable that is sampled from a probability distribution $\pi$, conditioned on the augmented latent state (i.e., $\dact_k \sim \policy(\dact_k | \s_k, \act_k))$. Even though the model is deterministic, we define the policy to be stochastic to allow for stochastic exploration. The policy will be specified further in \sectionref{sec:control}.
\begin{align}
    \dact_k &\sim \policy(\dact_k | \s_k, \act_k) \label{eq:policy}
\end{align}
\subsubsection{Decoder} After learning the latent model we intend to plan over it, which involves a multi-step prediction. Given only the encoder \eqref{eq:encoder}, dynamics model \eqref{eq:dynamics_model}, policy \eqref{eq:policy} and the current $(\obs_{k}, \act_{k})$-pair, we would be limited to single-step predictions of $\s_{k+1}$ at run-time, because multi-step predictions $\s_{k+i}$ with $i>1$ would require knowledge of future observations $\obs_{k+i}$ to evaluate $B_{\eig_{k+i}}$. Therefore, to make multi-step predictions, we introduce a decoder $\eig^\inv$ (parametrized by a NN) in \eqref{eq:decoder} that uses predicted latent states $\s_{k+i}$ to construct pseudo-observations $\obsh_{k+i}$ that produce the same partial derivative as the true observation (i.e., $\pde[\act_{k}]\eig(\eig^\inv(\s_k), \act_{k}) \approx \pde[\act_{k}]\eig(\obs_{k}, \act_{k})$). Future values $\act_{k+i}$ do not pose a problem, because they can be inferred from the policy \eqref{eq:policy} and dynamics model \eqref{eq:dynamics_model}. 
\begin{align}
	\obsh_k &= \eig^\inv (\s_k). \label{eq:decoder}
\end{align}
\subsubsection{Image processor}When the observations are raw pixel images $\pixel_k$, not all relevant information can be inferred from a single observation. To restore the Markov property, we pass the last $d$ consecutive pixel images through a convolutional neural network $\Omega$ in \eqref{eq:image_processor}, stack the output into a single vector, and consider that to be the observation $\obs_k$ instead. In that case, the observed sequences consist of tuples $t_k=(\pixel_k,...,\pixel_{k-d+1}, \act_k, \dact_k, \cost_k)$.
\begin{align}
	\obs_k &= \Omega(\pixel_k,...,\pixel_{k-d+1}),\label{eq:image_processor}
\end{align}
\subsection{Learning the latent model}\label{sec:training}
Our latent model should have linear dynamics and be predictive of observed costs. These two high-level requirements lead to the following three losses which are minimized during training.

\subsubsection{Linear dynamics} To ensure that the latent state is a valid Koopman eigenfunction, we regularize the time evolution in the latent space to be linear by using the following loss,
\begin{align}
	&\text{Linear loss:}	&\loss{lin} &= \frac{1}{T} \sum^{T-1}_{k=0}\norm{\eig(\obs_{k+1},\act_{k+1})-\s_{k+1}}_\text{MSE}, \label{eq:loss_lin}
\end{align}
where $\s_{k+1}$ is obtained by rolling out a latent trajectory as illustrated in \figureref{fig:architecture}.

\subsubsection{Cost prediction} We want the latent representation to contain all necessary information to solve the task. If we would naively apply an autoencoder that predicts future observations, we focus the majority of the model capacity on potentially task-irrelevant dynamics contained in the observations. To learn a latent representation that \textit{only} encodes relevant information, we propose to use a lossy autoencoder that is predictive of current and future costs instead. Such a representation would allow an agent to predict the cost evolution of various action sequences and choose the sequence that minimizes the predicted cumulative cost, which is essentially equivalent to solving the task. Because we only penalize inaccurate cost predictions, the encoder is not incentivized to encode task-irrelevant dynamics into the latent representation as they are not predictive of the cost. This leads to the \bhr{task-relevant}{robust} identification of the lifting function $\eig$. Cost prediction accuracy is achieved by using the following two losses,
\begin{align}
	&\text{Reconstruction loss:}	&\loss{recon} &= \norm{\cost_0-\hat{\cost}_0}_\text{MSE} \label{eq:loss_recon} \\
	&\text{Prediction loss:}		&\loss{pred} &= \frac{1}{T} \sum^{T}_{k=1} \norm{\cost_{k}-\hat{\cost}_k}_\text{MSE} \label{eq:loss_pred}
\end{align}

\subsubsection{Training objective} We minimize the losses in \eqref{eq:loss_lin}, \eqref{eq:loss_recon}, and \eqref{eq:loss_pred}, corresponding to linear dynamics regularization and cost prediction, together with an $L2$-regularization loss $\loss{reg}$ on the trainable variables (excluding neural network biases). This leads to the following training objective,
\begin{align}\label{eq:objective_representation}
	\min_{\theta, \lambda^{[1],...,[P]}_{\pm}} \loss{lin} + \alpha_1(\loss{recon} + \loss{pred}) + \alpha_2\loss{reg},
\end{align}
where $\theta$ is the collection of all the trainable variables that parametrize the encoder $\eig$, decoder $\eig^\inv$, cost model $\psi$, and convolutional network $\Omega$ (in case of image observations). Weights $\alpha_1,\alpha_2$ are hyperparameters. The model is trained using the Adam optimizer \citep{Kingma2014a} with learning rate $\tilde{\alpha}$, on batches of $B$ sequences $\{t_k\}^{T}_{k=0}$ for $E$ epochs.

%% file: figures/network_pde.pdf_tex
\begingroup%
  \makeatletter%
  \providecommand\color[2][]{%
    \errmessage{(Inkscape) Color is used for the text in Inkscape, but the package 'color.sty' is not loaded}%
    \renewcommand\color[2][]{}%
  }%
  \providecommand\transparent[1]{%
    \errmessage{(Inkscape) Transparency is used (non-zero) for the text in Inkscape, but the package 'transparent.sty' is not loaded}%
    \renewcommand\transparent[1]{}%
  }%
  \providecommand\rotatebox[2]{#2}%
  \newcommand*\fsize{\dimexpr\f@size pt\relax}%
  \newcommand*\lineheight[1]{\fontsize{\fsize}{#1\fsize}\selectfont}%
  \ifx\svgwidth\undefined%
    \setlength{\unitlength}{531.11980288bp}%
    \ifx\svgscale\undefined%
      \relax%
    \else%
      \setlength{\unitlength}{\unitlength * \real{\svgscale}}%
    \fi%
  \else%
    \setlength{\unitlength}{\svgwidth}%
  \fi%
  \global\let\svgwidth\undefined%
  \global\let\svgscale\undefined%
  \makeatother%
  \begin{picture}(1,0.49965128)%
    \lineheight{1}%
    \setlength\tabcolsep{0pt}%
    \put(0,0){\includegraphics[width=\unitlength,page=1]{network_pde.pdf}}%
    \put(0.74596669,0.30314674){\color[rgb]{0,0,0}\makebox(0,0)[lt]{\lineheight{1.25}\smash{\begin{tabular}[t]{l}Train Model\end{tabular}}}}%
    \put(0.62448289,0.22331583){\color[rgb]{0,0,0}\rotatebox{90}{\makebox(0,0)[lt]{\lineheight{1.25}\smash{\begin{tabular}[t]{l}Buffer\end{tabular}}}}}%
    \put(0.55121252,0.03115832){\color[rgb]{0,0,0}\makebox(0,0)[lt]{\lineheight{1.25}\smash{\begin{tabular}[t]{l}\textbf{DeepKoCo}\end{tabular}}}}%
    \put(0.00863325,0.37669286){\color[rgb]{0,0,0}\makebox(0,0)[lt]{\lineheight{1.25}\smash{\begin{tabular}[t]{l}\textbf{Rollout}\end{tabular}}}}%
    \put(0.07336043,0.09086154){\color[rgb]{0,0,0}\makebox(0,0)[lt]{\lineheight{1.25}\smash{\begin{tabular}[t]{l}\textbf{Latent Model}\end{tabular}}}}%
    \put(0.89340464,0.2820741){\color[rgb]{0,0,0}\makebox(0,0)[t]{\lineheight{1.25}\smash{\begin{tabular}[t]{c}OU\\Noise\end{tabular}}}}%
    \put(0.87423688,0.1556291){\color[rgb]{0,0,0}\makebox(0,0)[lt]{\lineheight{1.25}\smash{\begin{tabular}[t]{l}MPC\end{tabular}}}}%
    \put(0,0){\includegraphics[width=\unitlength,page=2]{network_pde.pdf}}%
  \end{picture}%
\endgroup%

%% file: mpc.tex
\section{DEEP KOOPMAN CONTROL}\label{sec:control}
This section introduces the agent that uses the Koopman latent model to find the action sequence that minimizes the predicted cumulative cost. We use linear model-predictive control (LMPC) to allow the agent to adapt its plan based on new observations, meaning the agent re-plans at each step. Re-planning at each time-step can be computationally costly. In the following, we explain how to exploit and adapt our latent model and cost model to formulate a sparse and convex MPC problem that can be solved efficiently online.

The planning algorithm should achieve competitive performance, while only using a limited amount of computational resources. This motivates choosing Koopman eigenfunctions as the latent state, because the \emph{autonomous} dynamics are linear. The dynamics are affine in the controlled variable $\dact_k$ that is multiplied in the definition of the state space by $B_{\eig_k}$, which depends on the latent state. Similarly, $C_{\s_k}$ requires the evaluation of the nonlinear function $\psi(\s_k)$. There exist methods that can be applied in this setting, such as the State-Dependent Ricatti Equation (SDRE) method \citep{Chang2013a}. While the SDRE requires less complexity compared to sample-based nonlinear MPC (e.g. CEM \citep{Szita2006a}), it remains computationally demanding as it also requires the derivative of $\psi$ with respect to $s_k$ at every step of the planning horizon. \looseness=-1

Our goal is to reduce the online complexity of our planning strategy, while also dealing with input constraints. Hence, we trade-off some prediction accuracy (due to the mismatch between the latent model and the MPC prediction model) to simplify the online planning strategy by using linear MPC. We propose to evaluate the state-dependent matrices $C_{\s_0}$ and $B_{\eig_0}$ at time-step $k=0$ (obtained from our latent model) and keep them both fixed for the rest of the LMPC horizon. This assumes that the variation of $B_{\eig_k}$ and $C_{\s_k}$ is limited over the prediction horizon (compared to \eqref{eq:dynamics_model} and \eqref{eq:cost_model}). Nevertheless, thanks to this simplification we can rely on LMPC for planning that can be solved efficiently. Specifically, once we evaluate $\s_0$, $C_{\s_0}$, and $B_{\eig_0}$, the computational cost of solving the MPC problem in the \emph{dense form} \citep{Maciejowski2002a} scales linearly with the latent state dimension due to the diagonal structure of $\Lambda$. As Section \ref{sec:results} details, this simplification allows our method to achieve competitive final performance, while only requiring a single evaluation of the NNs $\Omega$, $\eig$, and $\psi$. This significantly decreases the computational cost at run-time compared to sample-based nonlinear MPC (e.g., CEM \citep{Szita2006a}) that would require many evaluations of the NNs at every time-step. In contrast to LQR, LMPC can explicitly deal with actuator saturation by incorporating constraints on $\act$.\bhr{}{ An overview of the proposed method is shown in \algorithmref{alg:deepKoCo}.} The proposed planning strategy based on LMPC is defined as follows:
\begin{align}
	\min_{\dact^{\idx{0}, \dots,\idx{H-1}}}  &\hspace{0.5em} \sum^{H}_{k=1} ||C_{\s_0}\s_k||^2_2 + \act^\tr_k R \act_k + \dact^\tr_k \tilde{R} \dact_k, \label{eq:objective_MPC}\\
	\mathrm{s.t.}   &\hspace{0.5em} \begin{bmatrix} \s_{k+1} \\ \act_{k+1} \end{bmatrix} = \begin{bmatrix} \Lambda & 0 \\ 0 & I \end{bmatrix} \begin{bmatrix} \s_k \\ \act_k \end{bmatrix} + \begin{bmatrix} B_{\eig_0} \\ I \end{bmatrix} \dact_k,  \nonumber \\
	& \hspace{0.5em} \act^{\min} \leq \act_{k} \leq \act^{\max} \text{, for } k=1,\dots, H, \nonumber
\end{align}
where $H$ is the prediction horizon. Positive-definite matrix $\tilde{R}$ penalizes the use of $\dact_k$ and is required to make the problem well-conditioned. Its use does introduce a discrepancy between the approximate cost model \eqref{eq:cost_model} and the cumulative cost ultimately minimized by the agent \eqref{eq:objective_MPC}. Therefore, the elements in $\tilde{R}$ are kept as low as possible.

To align the representation learning objective \eqref{eq:objective_representation} with the linear MPC objective \eqref{eq:objective_MPC}, we also fix the state-dependent terms $C_{\s_0}$ and $B_{\eig_0}$ at time-step $k=0$ in the evaluation of the cost prediction loss \eqref{eq:loss_pred} and linear loss \eqref{eq:loss_lin}. Note that this does not mean that $C_{\s_k}$ and $B_{\eig_k}$ are constant in the latent model \eqref{eq:dynamics_model}, \eqref{eq:cost_model}. The matrices remain state-dependent, but their variation is \bhr{limited}{regularized} over the sequence length $T$. In general, we choose the sequence length to be equal to the prediction horizon.\bhr{ Hence, we learn a representation that provides local linear models that are particularly accurate around $\s_k$ in the direction of the (goal-directed) trajectories gathered during training.}{}

To gather a rich set of episodes to learn the Koopman latent model, we add colored noise to the actions commanded by the agent's linear MPC policy, that is, $\act_{k+1}=\act_k + \dact_k + \OU_k$. This adds a stochastic exploration component to the policy. We use an Ornstein-Uhlenbeck (OU) process to generate the additive colored noise with decay rate $\lambda^\text{ou}$. The variance $\sigma^{2, \text{ou}}$ is linearly annealed from $\sigma_\text{init}^{2, \text{ou}}\rightarrow 0$ over $1,\dots,N^\text{ou}$ episodes, that is, after $N^\text{ou}$ episodes the policy becomes deterministic.\looseness=-1

\bhr{An overview of the proposed method is shown in \algorithmref{alg:deepKoCo}. First, we initialize all model parameters. Then, we construct the Koopman operator $\Lambda$ with \eqref{eq:jordan_block} and gather $N$ episodes of experience. Each time-step, we process the image observations with \eqref{eq:image_processor}, evaluate the latent model \eqref{eq:encoder}, \eqref{eq:dynamics_model}, and \eqref{eq:cost_model}, and use it to find $\dact_k$ with \eqref{eq:objective_MPC}. Noise is added to the action increment before it is applied to the environment. We fill the experience buffer $D$ with $N$ episodes, split into sequences of length $T$, and train on them for $E$ epochs with \eqref{eq:objective_representation}. This is repeated until convergence.}{}

{\begin{algorithm}[t]
\fontsize{10}{10}\selectfont
	\caption{Deep Koopman Control (DeepKoCo)\bhr{}{. Function \textit{GetKoopmanOperator} constructs $\Lambda$ with \eqref{eq:jordan_block}, \textit{ProcessImages} implements \eqref{eq:image_processor}, \textit{LatentModel} implements \eqref{eq:encoder}, \eqref{eq:dynamics_model}, and \eqref{eq:cost_model}, \textit{LMPC} implements \eqref{eq:objective_MPC}, and \textit{TrainModel} minimizes \eqref{eq:objective_representation}. The remaining functions have names that describe their high-level purpose.}}
	\label{alg:deepKoCo}
	\DontPrintSemicolon
	\LinesNumbered
	\KwIn{Model parameters: $P, d$ 
	\\ \quad \qquad Policy parameters: $\zeta=\{H, R, \tilde{R}\}$
	\\ \quad \qquad Noise parameters: $\lambda^\text{ou}, \sigma_\text{init}^{2, \text{ou}}, N^\text{ou}$
	\\ \quad \qquad Train parameters: $N, L, T, E, B$
	\\ \quad \qquad Optimization parameters: $\xi=\{\alpha_1, \alpha_2, \tilde{\alpha}\}$
	}
	\KwOut{Eigenvalues $\lambda^{[1],...,[P]}_{\pm}$
	\\ \qquad \qquad Trained networks $\eig, \eig^\inv, \psi, \Omega$}
	$\theta,\lambda^{[1],...,[P]}_{\pm}\leftarrow$ InitializeModel$(P, R)$\;
	\While{not converged}{
		$\Lambda \leftarrow$ GetKoopmanOperator$(\lambda^{[1],...,[P]}_{\pm})$ \;
		\For{episode $l=1,\dots,N$}{
			$\pixel_0,...,\pixel_{1-d}\leftarrow$ ResetEnvironment() \;
			$\act_{0}\leftarrow$ 0 \;
			\For{time-step $k=0,\dots,L$}{
			    $\obs_k \leftarrow$ ProcessImages$(\pixel_k,...,\pixel_{k-d+1}, \theta)$ \;
			    $\s_k, B_{\eig_k}, C_{\s_k} \leftarrow$ LatentModel$(\obs_{k}, \act_{k}, \theta)$ \;
				$\dact_k \leftarrow$ LMPC$(\s_k, \act_k, B_{\eig_k}, C_{\s_k}, \Lambda, \zeta)$ \;
				$\act_{k+1} \leftarrow \act_{k} + \dact_k + $ Noise$(\lambda^\text{ou}, \sigma^{2, \text{ou}})$\;
				$\pixel_{k+1}, \cost_{k}\leftarrow$ ApplyAction$(\act_{k})$ \;
			}
			$D \leftarrow D \cup$ CreateSequences$(T,\{t_k\}^{L}_{k=0})$
		}
		$\theta,\lambda^{[1],...,[P]}_{\pm}\leftarrow$ TrainModel$(D,\theta,\lambda^{[1],...,[P]}_{\pm},\xi,E, B)$ 
	}
\end{algorithm}}

%% file: results.tex
\begin{figure*}[t]
    \centering
    \includegraphics[width=\textwidth]{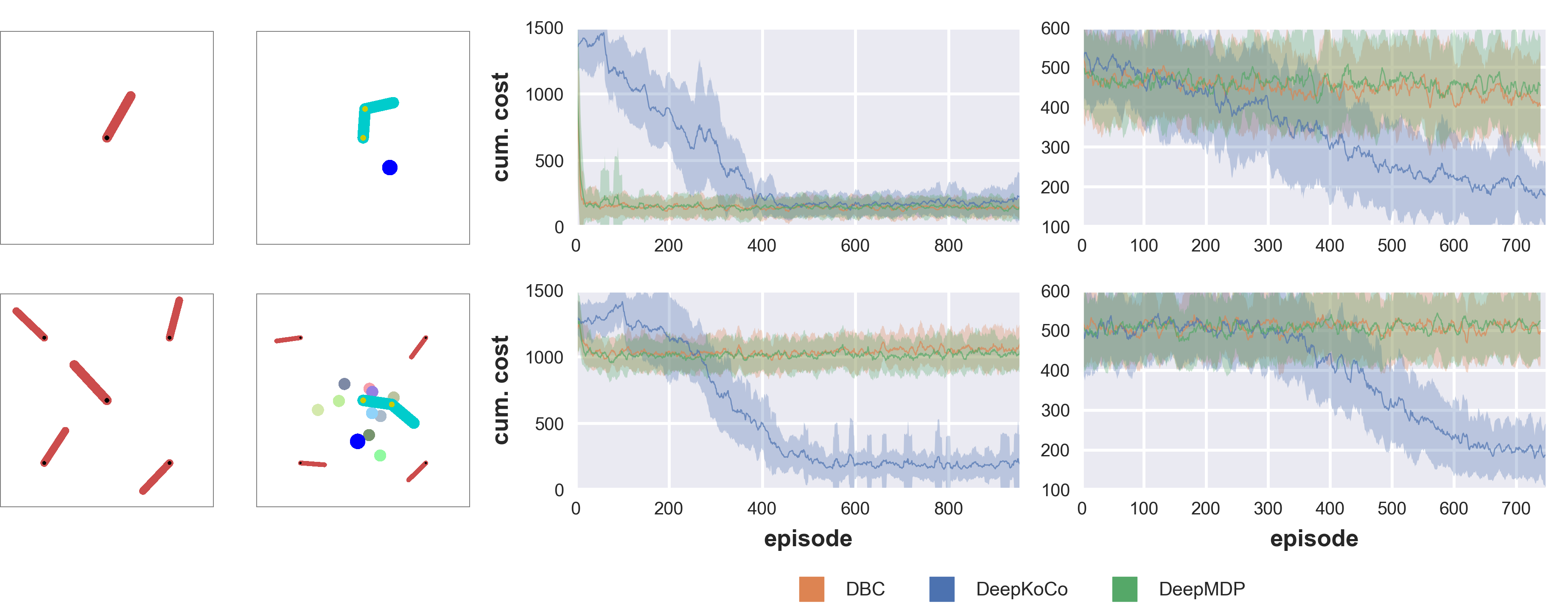}
    \caption{\textbf{Left} Typical setups for the two tasks in a clean scenario (first row) and distractor scenario (second row). In all setups, the center system is controlled by the agent. In the manipulator task, the moving target is a blue ball. \textbf{Right} Learning curves when using state observations. The grid-location of each figure corresponds to the grid-location of each setup on the left. The mean cumulative cost over the last 10 episodes (line) with one standard deviation (shaded area) over 5 random seed runs are shown.\vspace*{-4mm}}\label{fig:results}
\end{figure*}

\begin{figure*}[t]
    \centering
    \includegraphics[width=\textwidth]{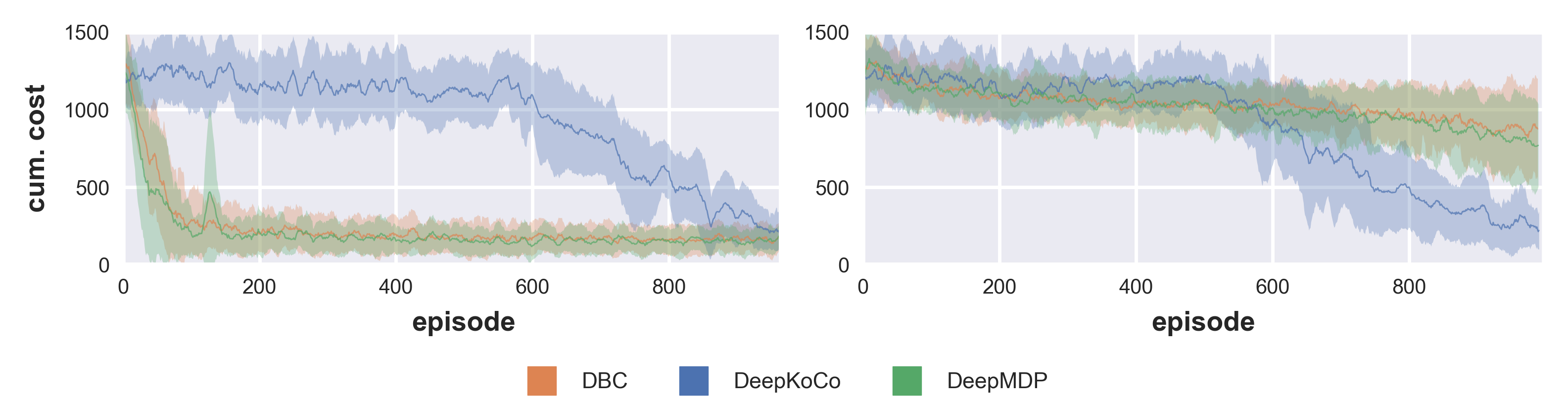}
    \caption{Learning curves when using images as observations in the pendulum task. The mean cumulative cost over the last 10 episodes (line) with one standard deviation (shaded area) over 5 random seed runs are shown. \textbf{Left} Clean scenario. \textbf{Right} Distractor scenario.\vspace*{-4mm}}\label{fig:results_images}
\end{figure*}

\section{RESULTS}
\label{sec:results}
We evaluate \textit{DeepKoCo} on two continuous control tasks, namely OpenAI's pendulum swing-up task and a manipulator task. The manipulator task is similar to OpenAI's reacher task where the two joints of a two-link arm are torque controlled and the Euclidean distance between the arm's end-point and a target must be minimized. However, we increase the difficulty by allowing the target to move at a constant angular velocity and radius around the arm's center-joint. Given that the angular velocity and radius vary randomly over episodes, the manipulator must learn to track arbitrary circular trajectories at different speeds. The dynamics for the manipulator can be formulated as $\nls_{k+1} = F(\nls_k) + B(\nls_k) \act_k$, where $\nls$ is the original nonlinear state. Such dynamics do not necessarily admit a Koopman representation that is (bi-)linear in the states and actions, as is often assumed in literature \cite{Li2020a, Mamakoukas2019}. 

To investigate the effect of distractor dynamics, we test each task in two different scenarios. In the first scenario, only relevant dynamics are observed, while in the second one we purposely contaminate the observations with distractor dynamics. The agent is either provided with a concatenated state observation containing the state measurements of both the relevant system and distractors \bhr{(while not knowing which states are the relevant ones)}{} or image observations with all systems in-frame (refer to \figureref{fig:results} for the setup). The state observation dimension from the clean to the distractor scenario increases from $3$ to $15$ for the pendulum and from $10$ to $50$ for the manipulator. A video of the simulations using DeepKoCo accompanies the paper~\citep{DeepKoCoVideo}.

\subsubsection{Baselines}We compare with two baselines that both combine model-free reinforcement learning with an auxiliary bisimulation loss to be robust against task-irrelevant dynamics: \emph{(i)} Deep Bisimulation for Control (DBC) \citep{Zhang2020a}, and \emph{(ii)} DeepMDP \citep{Gelada2019a}. In case of state observations, we replace their convolutional encoder, with our fully connected encoder.

\subsubsection{Hyperparameters}We use the same set of hyperparameters throughout all experiments, except for the number of complex eigenvalue pairs $P$, that is, $P=10$ and $P=30$ in the swing-up and manipulator task, respectively to cope with the complexity of the scenarios. As policy parameters, we use $H=15, R=0.001, \tilde{R}=0.01, \lambda^\text{OU}=0.85, \sigma_\text{init}^{2, \text{ou}}=0.85$. Initially, we fill the experience buffer $D$ with $N=90$ episodes, split into sequences of length $T=15$, and train on them for $E=100$ epochs. Then, we continuously add the sequences of $N=20$ episodes to the buffer and train on the complete buffer for $E=3$ epochs. As optimization parameters, we use $\alpha_1=10, \alpha_2=10^{-14}, \tilde{\alpha}=0.001$. In case of image observations, we stack the last $d=3$ images, downsample them to $3\times64\times64$ pixels before passing them through the convolutional NN defined in \citet{Ha2018a}. The networks $\eig, \eig^\inv, \psi$ are 2-layered fully connected NNs with 90, 90, and 70 units per layer, respectively. The layers use ReLU activation and are followed by a linear layer. \bhr{The number of complex eigenvalue pairs $P$, planning horizon $H$, and action increment cost $\tilde{R}$ are the most important parameters to tune.}{}

\subsubsection{Clean scenario} Concerning the pendulum task, the baselines converge more quickly to the final performance compared to DeepKoCo, as the top-left graph of \figureref{fig:results} shows. Nevertheless, we do consistently achieve a similar final performance. The slower convergence can be explained by the added noise, required for exploration, that is only fully annealed after 400 episodes. We believe the convergence rate can be significantly improved by performing a parameter search together with a faster annealing rate, but we do not expect to be able to match the baselines in this ideal scenario. Note that, despite the apparent \emph{simplicity} of the application scenario, finding an accurate Koopman representation for the pendulum system is challenging, because it exhibits a continuous eigenvalue spectrum and has multiple (unstable) fixed points \citep{Lusch2018}. Concerning the manipulator task, both baselines were not able to solve the manipulator task with a moving target as the top-right graph of \figureref{fig:results} shows, while they were able to learn in case the target was fixed (results omitted due to space limitations). This shows that learning to track arbitrary circular references is significantly harder than regulating towards a fixed goal. Despite the increased difficulty, DeepKoCo learns to track the moving target. Finally, note that the manipulator task shows that the proposed method can deal with a multi-dimensional action-space and non-quadratic cost functions, that is, the Euclidean norm (the square root of an inner product).

\subsubsection{Distractor scenario} In the more realistic scenario, both baselines fail to learn anything. In contrast, our approach is able to reach the same final performance as in the clean scenario, in a comparable amount of episodes, as the bottom row of \figureref{fig:results} shows. This result can be explained by noticing that our multi-step cost prediction (in our loss function \eqref{eq:loss_pred}) provides a stronger learning signal for the agent to distinguish relevant from irrelevant dynamics. For the manipulator task, there is a tracking error caused by the trade-off of using fixed $B_\eig$ and $C_s$ along the MPC prediction horizon for efficiency. While our latent model presented in \sectionref{sec:learning_invariant_koopman_eigenfunctions} supports state-dependent matrices, we decided to keep them fixed in the control design for efficiency. 

\subsubsection{Image observations} \figureref{fig:results_images} shows the results for the pendulum task when images are used instead of state observations. In both clean and distractor scenarios, our approach is able to reach a similar final performance compared to using state observations. As expected, the baselines struggle to learn in the distractor scenario. This supports our statement that our approach learns a \bhr{task-relevant }{}Koopman representation from high-dimensional observations\bhr{}{ that is robust to task-irrelevant dynamics}. We plan to test the manipulator task with images both in simulation and in real-world experiments.

%% file: future_work.tex
\section{CONCLUSIONS AND FUTURE WORK}
We presented a model-based agent that uses the Koopman framework to learn a latent Koopman representation from images. DeepKoCo can find Koopman representations that \emph{(i)} enable efficient linear control, \emph{(ii)} are robust to distractor dynamics. Thanks to these features, DeepKoCo outperforms the baselines (two state-of-the-art model-free agents) in the presence of distractions. As part of our future work, we will extend our deterministic latent model with stochastic components to deal with partial observability and aleatoric uncertainty. Furthermore, we will extend our cost model to deal with sparse rewards, as they are often easier to provide.